%% file: naacl2021.tex
\documentclass[11pt]{article}

\usepackage[]{naacl2021}

\usepackage{times}
\usepackage{latexsym}

\usepackage[T1]{fontenc}

\usepackage[utf8]{inputenc}

\usepackage{microtype}
\usepackage{graphicx}

\newcounter{dlnr}
\newcommand{\dl}{\addtocounter{dlnr}{01}(\thedlnr)}

\newenvironment{dialogue-lu}%
   {\begin{it}%
    \begin{list}{}%
         {\setlength{\leftmargin}{.6cm}}%
         \item[]}%
   {\end{list}%
    \end{it}} 

\title{A recipe for annotating grounded clarifications}

\author{Luciana Benotti \\
  Universidad Nacional de C\'ordoba \\
  CONICET, Argentina \\
  \texttt{luciana.benotti@unc.edu.ar} \\\And
  Patrick Blackburn \\
Philosophy and Science Studies\\
IKH, Roskilde University, Denmark \\
  \texttt{patrickb@ruc.dk} \\}

\begin{document}
\maketitle
\begin{abstract}

In order to interpret the \emph{communicative intents} of an utterance, it needs to be \emph{grounded} in something that is \emph{outside of language}; that is, \emph{grounded} in  \emph{world modalities}. In this paper we argue that dialogue clarification mechanisms make explicit the process of interpreting the communicative intents of the speaker's utterances by grounding them  in the various modalities in which the dialogue is situated. 
This paper frames dialogue clarification mechanisms as an understudied research problem and a key missing piece in the giant 
jigsaw puzzle of natural language understanding. We discuss both the theoretical background and practical challenges posed by this problem, and propose a recipe for obtaining grounding annotations. 
We conclude by highlighting ethical issues that need to be addressed in future work. 

\end{abstract}

\input{introduction}

\input{theoretical-background}

\input{recipe}

\input{annotation} 

\input{analysis}

\input{discussion}

\input{conclusions}

\section*{Acknowledgments}

We thank the anonymous reviewers for their detailed reviews and insightful comments.

\input{ethical_concers}

\bibliographystyle{acl_natbib}
\bibliography{benotti,eacl2021}

\appendix

\input{supplementary.tex}

\end{document}

%% file: introduction.tex
\section{Introduction}

Clarifications are crucial to robust dialogues, and pragmatic factors ---
notably those shaped by the world modalities situating the conversation  ---
have a key role to play. \emph{Referring expressions} have in \emph{vision} a modality in which to ground clarifications concerning objects in the world~\cite{vries:2017}; \emph{navigation instructions} have in \emph{movement} a modality in which to ground clarifications concerning collaborative wayfinding~\cite{thomason:2019}. 
Clarifications grounded in situationally relevant modalities
boost the \emph{redundancy} required to learn to use language 
without explicit supervision, as they make explicit the process of negotiating the \emph{communicative intent}. But despite its importance, work on clarification remains scattered.

Humans switch between clarifications grounded in different modalities seamlessly but
(we shall argue) systematically. Our discussion is based
around a general recipe for detecting \emph{grounded clarifications};
we work towards this in  Section~\ref{theoretical-background} by first reviewing the distinction between perceptual and collaborative grounding, and then discussing clarification mechanisms, Clark~\shortcite{clark96}'s action ladder of communication, and  Ginzburg, Purver and colleagues~\shortcite{ginzburg:2012}'s classification of clarification phenomena. In Section~\ref{CR-id} we draw these threads together and present the central idea:
\begin{quote}
\textit{Given an utterance U, a subsequent turn is its clarification grounded in modality \emph{m} if it cannot be preceded by positive evidence of understanding of U in \emph{m}}. 
\end{quote}
This provides a unified way to frame clarification mechanisms and their interactions across various modalities;
a graphical specification of the recipe it gives rise to can
be found in Figure~2 of the supplementary material.
It covers clarifications grounded in moving, grabbing and changing the physical world: these have traditionally been considered plain-old-questions~\cite{purver:2018}, but we view them as
useful clarification ingredients.%
\footnote{We are suspicious of the common assumption that requests for information regarding references that are grounded in vision (e.g. \textit{the red or the blue jacket?}) are clarifications, whereas requests for information grounded in other modalities are not (e.g. \textit{do I take the stairs up or down?}).
} 
In Sections~\ref{sec-corpus-4} and~\ref{sec-corpus} 
we test the practical implications of our recipe by identifying and characterizing (according to their modalities) the clarifications in a corpus of long dialogues in English.
In Section~\ref{sec-to-CR-or-not} we turn to the
claim that clarifications are rare in dialogue datasets~\cite{ginzburg:2012}, and that current data-hungry algorithms cannot learn them. 
We argue that whether they are rare or not depends on pragmatic factors of the conversation and the modality of the grounded clarification, 
and 
discuss the impact of six such factors. After presenting potential objections and our responses in
Section~\ref{discussion}, we conclude in Section~\ref{sec-conclusions-cap2}
by noting 
ethical issues raised by
socioperceptive dialogue systems that will need to be addressed.%
\footnote{See also the supplement on ethical considerations.}



%% file: theoretical-background.tex
\section{Theoretical Background} \label{theoretical-background}

We begin by reviewing the theoretical background on grounding and clarification mechanisms. We then
examine two schemes proposed to characterize clarifications according to their conversational function: one focuses on the problem of anchoring utterance parameters into the conversational history,   the other emphasizes a multimodal ladder of actions co-temporal with dialogue turn-taking.
We are interested in the potential contributions of both
towards a recipe for annotating clarification mechanisms.

\subsection{Collaborative and perceptual grounding}
\label{previous}

\emph{Collaborative grounding} is the process of seeking and providing incremental evidence of mutual understanding through dialogue. 
When the speaker believes that the dialogue is on track, \emph{positive evidence} of understanding is provided in different forms (depending on the channel of communication) such as explicit acknowledgements, and via backchannels such as nods, eye contact, etc. \emph{Negative evidence} of understanding signals that something needs negotiation before the dialogue partners can commit;
 clarification requests are the prototypical example of negative evidence.  

Collaborative grounding is distinct from perceptual (or symbol) grounding~\cite{harnad:1990,He:2016,bansal:2019, Lu:2020}. The perceptual grounding literature deals with capabilities enabling symbols to be linked with perceptions, and is rooted in situationally
relevant modalities such as vision. Collaborative grounding, on the other hand, deals with the 
\emph{dynamics  of conversation}
(the  ongoing exchange of speaker and hearer roles)
and is rooted in situationally relevant aspects of
socioperception.
Alikhani and Stone~\shortcite{alikhani:2020} note several basic mechanisms that contribute to collaborative grounding, including those for dealing with joint attention~\cite{koller-etal-2012-enhancing,koleva-etal-2015-impact,bohus:2020}, engagement~\cite{bohus:2014,foster:2017}, turn taking and incremental interpretation~\cite{schlangen-skantze-2009-general,selfridge:2012,devault-traum-2013-method,eshghi:2015} corrections and clarifications~\cite{villalba-etal-2017-generating,ginzburg-fernandez} and dialogue management~\cite{devault-stone-2009-learning,selfridge:2012}. These mechanisms have been studied for different kinds of applications~\cite{denis:2010,dzikovska:2010,dzikovska:2012}. 
Both collaborative and perceptual
grounding are important (\textit{all} relevant
modalities are potentially important) and
in this paper we bring them together under an umbrella we call \textit{grounded clarification}.

\subsection{Clarification mechanisms}
\label{sec-CR-role}

Clarification requests (CRs) and their answers 
are the prototypical clarification mechanisms (CMs), pieces
of dialogue that participants
use to signal lack of understanding and to trigger
negotiation. CMs are used in all kinds of dialogue and  are influenced by the type of interaction, the dialogue participants, and the context in which the conversation occurs. 
Interest in CMs by the artificial intelligence community dates back to the start of the century, and has typically focused on mechanisms for human-computer dialogue systems~\citep{gabsdil03,purver04,rodriguez04,rieser05,skantze07}. In sociolinguistics and discourse analysis, on the other hand, the interest in CMs (or \textit{repairs}, as they are usually called there) has focused on human-human conversation for over three decades now; see~\citep{schegloff87b} for a representative example.

How CMs can be learned from data remains understudied. Rao and Daum{\'e}~III~\shortcite{rao2018} rank clarification requests of stackoverflow articles according to their usefulness: a good clarification question is one whose expected answer will be useful, 
which means that the clarification highlighted important information missing from the initial request for help; we share this view, but differ from Rao and Daum{\'e}~III, in that we focus on CMs and their responses occuring in multiturn dialogue. 

It may seem plausible to expect that clarification requests will be realized as questions; however, corpus studies indicate that their most frequent realization  is in declarative form~\cite{jurafsky04}. Indeed, the form of a clarification request~\cite{rodriguez04} is \textit{not} a reliable indicator of the function that the clarification request is playing. Neither does form unambiguously indicate whether a dialogue contribution is a CR or not. The surface form of explicit negotiations of meaning in  dialogue are frequently non-sentential utterances~\cite{fernandez:2006,fernandez:2007}. These include the prototypical positive and negative evidence of grounding (acknowledgements and clarification requests~\cite{stoyanchev:2013}) but also less-well-known forms such as self-corrections, rejections, and modifiers~\cite{purver04,purver:2018}.  These observations indicate that we face significant challenges if we want to train a system to seek or supply clarification  effectively.

\subsection{Clarifications grounded in parameters}
\label{function-schemes}

 Ginzburg, Purver and colleagues (henceforth G\&P)
 proposed the first scheme to classify the functions of CRs;
 see~\cite{purver03,purver06,ginzburg:2012}. The
G\&P classification uses the categories shown on
Table~\ref{GP_schema}.
The idea driving this work is that CRs are caused by problems
arising during the anchoring of  utterance parameters  into the
conversational history. 

\begin{table}[h]
 \begin{center}
\begin{small}
\begin{tabular}{ p{1.5cm}  p{2.5cm}  p{2.5cm} } 
  \hline
  \textsc{Category} & \textsc{Obstacle} & \textsc{Examples} \\
  \hline
  \emph{Repetition} & Cannot identify a surface 
parameter & What did you say?\\
  \hline
   \emph{Clausal}
  & Uncertain value for a clausal dialogue history
parameter & Are you asking if BO SMITH left?\\
  \hline
  \emph{Intended} & The hearer can find no value for a
 parameter
& Who is Bo? \\
  \hline
   \emph{Correction} & The hearer thinks that the speaker made a mistake and offers an alternative realization & Did you mean to say `Bro'?\\
   \hline
\end{tabular}
\end{small}
\end{center}
\caption[]{\label{GP_schema} CR classification scheme by P\&G}
\end{table}

The G\&P classification has been criticized~\citep{rodriguez04,rieser05}
because,
in practice, it seems difficult to decide what the category of a particular CR
is;
that is, CRs are usually ambiguous in this classification. 
In fact, G\&P recognize this issue themselves, pointing out that CRs that do not
repeat (part of) the content of the \emph{source utterance}
(that is, the utterance that is being
clarified) can exhibit all three readings. 

However, G\&P's classification is only ambiguous if \textit{only the past, but not the future, conversational history is taken into account}. 
It is crucial to analyze the CR \textit{response} in order to disambiguate the CR category. Sometimes the immediate linguistic context gives the clue necessary for disambiguation: whereas a repetition reading permits the responder to the CR to repeat her utterance verbatim, a clausal confirmation usually receives a yes/no answer, and an intended content reading requires the responder to reformulate in some way. Hence, the turn of the responder (and
the subsequent reaction of the participant originally making the CR) can disambiguate among readings. Consider the following example
from~\citep{purver04}. 
The example shows a case where George's initial clausal interpretation is
incorrect (the initiator is not satisfied), and a constituent reading
is required (Anon cannot find a value for Spunyarn).

\begin{dialogue-lu}
George: you always had er er say every foot he had with a piece of
spunyarn in the wire \\
Anon: Spunyarn? \\
George: Spunyarn, yes \\
Anon: What's spunyarn? \\
George: Well that’s like er tarred rope
\end{dialogue-lu}

In other situations, the immediate linguistic context will not be enough
(for instance, a reformulation can be a good response to all three types of CRs)
and then the whole conversational history might need to be analyzed in order to
disambiguate. This makes G\&P's classification difficult to use in
annotation studies where the annotators only get shallow, partial, localized
views of the dialogues.

\subsection{Clarifications grounded in modalities}
\label{act-schemes}

The second classification we shall examine puts the conversational action modality in the central role; it has been used in formal approaches to handling clarifications in dialogue systems~\citep{gabsdil03,rodriguez04,rieser05}. This classification is based on the four-level model of conversational action independently developed
by~\cite{allwood95} and~\cite{clark96}. Here, we use
Clark's terminology; his model is reproduced in Table~\ref{levels}. 

\begin{table}[h]
 \begin{center}
\begin{small}
\begin{tabular}{ l  l  l } 
  \hline
   L & \textsc{Speaker A's actions} & \textsc{Addressee B's
actions} \\
  \hline
   4 & \emph{Propose} project w to B & \emph{Uptake} A's
proposal w\\
  \hline
  3 & \emph{Intend} that B does i & \emph{Recognize} i from A
\\
  \hline
  2 & \emph{Present} signal s to B & \emph{Perceive} signal s
from A\\
  \hline
  1  & \emph{Execute} behavior t for B & \emph{Attend} to
behavior t from A \\
  \hline
\end{tabular}
\end{small}
\end{center}
\caption{\label{levels} Ladder of actions involved in communication}
\end{table}

Clark proposed this model in order to move from Austin's controversial
classification%
\footnote{For discussion of the controversies around Austin's classification of speech acts see~\cite{clark96}} of speech acts~\citep{austin62} to
a \emph{ladder of actions} which characterizes
not only the actions that are performed in language use (as Austin's does) but
also their inter-relationships. Clark~\shortcite{clark96} defines a ladder of actions as a set of co-temporal actions which provide
\textit{upward causality} and \textit{downward evidence}. Let us discuss these
using Table~\ref{levels}; we will call the speaker
Anna and the addressee Barny. Suppose that Anna tells Barny to sit down. We might say that Anna is performing just one action: asking Barny to sit down. But it is easy to argue that she is performing four distinct, though co-temporal, actions
--- actions beginning and ending simultaneously. These actions are in a \textit{causal} relation going \textit{up} the ladder (from level 1 up to level 4): Anna must get Barny to attend her behavior t (level 1) \emph{in order to} get him to hear the words she is presenting in her signals (level 2). Anna must succeed at that \emph{in order to} get Barny to recognize what she means  (level 3), and she must succeed at that \emph{in order to} get Barny to
uptake the project she is proposing (level 4). In short, causality (do
something \emph{in
order to} get some result) climbs up the ladder; this property
Clark calls \emph{upward
causality}. 

The different levels are related to different human modalities.
We say that level~1 is grounded into \emph{socioperception},  an ability that humans developed for collaboration that is crucial for achieving joint attention~\cite{tomasello05}. Level~2 is grounded in \emph{hearing} if we use speech as our communication channel. Level~3 is grounded in \emph{vision} when it involves recognizing referents in the real world. 
Level~4 is grounded in \emph{kinesthetic} when it involves moving and acting in the real world. The classification, along with obstacles that the addressee may face in the various modalities during the interpretation of a conversational action, is shown in Table~\ref{CR-schlangen}. In the rest of the paper we will refer to these modalities using the level number. 
 
\begin{table}[h]
 \begin{center}
\begin{small}
\begin{tabular}{ l  l  p{4cm} } 
  \hline
   L & \textsc{Modality} & \textsc{Examples} \\
  \hline
  4 & Kinesthetic & Do I take the stairs up or down?\\
  \hline
  3 & Vision & The red or the blue jacket? \\
  \hline
  2 & Hearing & What did you say?  \\
    \hline
  1 & Socioperception & Are you talking to me? \\
  \hline
\end{tabular}
\end{small}
\end{center}
\caption{\label{CR-schlangen} Ladder of actions grounded in modalities.}
\end{table}

Humans systematically use the evidence
provided by this ladder. Observing Barny sitting down is good evidence that he did not refuse to uptake (level 4) but also recognized what Anna intended and identified the chair (level 3). That is also evidence that she got Barny to hear her words (level 2), and evidence that she got him to attend to her (level 1). That is, evidence trickles \emph{down} the ladder;  Clark calls this the \emph{downward evidence} property.

If Barny repeats verbatim what Anna said (e.g. suppose she spoke in Spanish and
he repeats the word \emph{sientate}), then Anna has good evidence that he heard what she said (level 2). However, that is not necessarily evidence that he has recognized her intention; there might be an obstacle in level 3 (for instance, Barny
might not know Spanish). If there is such an obstacle, she would have completed levels~1 and~2 while failing to complete not only level~3 but also level~4 (it is rather unlikely that Barny would sit down right after hearing Anna --- and even if he did, this would not be \emph{because} he was uptaking Anna's project). A high level
action in the ladder can only be completed by executing \textit{all} the actions in the lower levels. This property Clark calls \emph{upward
completion}. 

If you tell somebody something, you expect a reaction from him.
If he doesn't answer, you might think that he didn't hear you, that he doesn't want to answer, or that he thinks you are talking to somebody else. None of these situations is very agreeable; humans don't like wasting effort, or being ignored. In order not to annoy the speaker, the addressee has two options: either he shows evidence in level 4 (and then, by downward evidence, the speaker knows that all the levels succeeded), or he indicates the obstacle in executing the  action (in any level). 
Clarifications are the tools that addressees can use  to make the obstacle explicit.

%% file: recipe.tex
\section{A grounded clarification recipe} 
\label{CR-id}

In this section we draw these threads together under the heading \emph{grounded clarification}. First, what is a clarification? Our starting proposal, which  we will modify, is the following:
\textit{given an utterance U, a subsequent turn is its clarification if it cannot be preceded by positive evidence of U}.
Note that this proposal implicitly embodies a procedure for annotating clarifications, one which could be crowdsourced: \textit{Is this a clarification? Check whether it can be preceded by positive evidence!} 

Our starting proposal is a modified version of
Gabsdil~\shortcite{gabsdil03}'s  test for CRs. Gabsdil 
says that CRs (as opposed to other kinds of dialogue contributions)
cannot be
preceded by explicit acknowledgments.  For example:

\begin{dialogue-lu}
   Lara: There's only two people in the class. \\
  a) Matthew: Two people? \\
  b) (*) Matthew: Ok, Two people? \\
  (BNC, taken from~\citep{purver03})
\end{dialogue-lu}

Gabsdil argues that (a) in the example above \emph{is} a CR because (b) is odd (we mark odd turns with (*) in examples). 
In (b), Matthew first acknowledges Lara's turn and only then indicates that her turn
contains information that he finds controversial.\footnote{This could be a felicitous
response, but it would require marked intonation to induce
a backtracking effect.}

On the other hand, (b) in the example below is fine and hence (a) \emph{is not} a
CR: the lieutenant acknowledges the sergeant's turn and then moves on to address 
what has become the most pressing 
topic in the conversation:

\begin{it}
\medskip \noindent
  Sergeant: There was an accident sir \\
  a) Lieutenant: Who is hurt? \\
  b) Lieutenant: Ok. Who is hurt? \\
 Adapted from~\cite[p.391]{traum03}
\medskip 
\end{it}

However Gabsdil's original test incorrectly discards cases
that we view as CRs. Consider the following example: 

\begin{it}
\medskip \noindent
 G: I want you to go up the left hand side of it
towards the green bay and make it a slightly diagonal
line, towards, sloping to the right. \\
F: Ok. So you want me to go above the carpenter? \\
\smallskip
Adapted from~\cite[p.30]{gabsdil03}
\medskip 
\end{it}

The problem is that the level of positive evidence  contributed by F's acknowledgment is ambiguous.  For instance, the Ok could (conceivably) mean:
\begin{itemize} 
\setlength{\itemsep}{0cm}
 \setlength{\parskip}{0cm}
\item \emph{Ok, so you want to talk to me} (level 1). 
\item \emph{Ok, I heard you} (level 2).
\item \emph{Ok, I saw what you are referring to} (level 3).
\item \emph{Ok, I did it}  (level 4, the highest level).
\end{itemize}
Thus we modify Gabsdil's test to make it \textit{level-sensitive}. In order to signal that all the levels have been
successful and that no CR related to any of them is expected, the simple
acknowledgment needs to be replaced by positive evidence in the highest level. This works for
Gabsdil's example: 

\begin{it}
\medskip \noindent
G: I want you to go up the left hand side of it
towards the green bay and make it a slightly diagonal
line, towards, sloping to the right. \\
(*) F: Ok, I did it. So you want me to go above the carpenter? 
\medskip 
\end{it}

Here \emph{So you want me to go above the carpenter?} is either weird
or far more likely to be interpreted as a question about an action that comes
\emph{after} F has successfully followed G's instruction. That is: it could be interpreted as F taking the initiative and proposing the next move, rather than as clarifying G's instruction.  Whether this is plausible would be determined by the following turns. 

More generally, if the addressee wants to uptake the speaker's proposal then he or she has two options: either to give positive evidence at the highest modality (and then, by downward closure, the speaker knows that all lower levels
succeeded) or to explicitly indicate the problem using a clarification (at any level). Table~\ref{CR-schlangen} illustrates, for each level and modality, possible CRs. We are not exhaustive about all the modalities that could happen in reality. We list four of them here but there could be more depending on the task.  

This approach to CR identification and classification is useful not only for instructions but also for other types of utterances. The following is an extension of Grice's classic implicature example (physical actions are between square brackets): 

\begin{it}
\medskip \noindent
A: I am out of petrol.\\
B: There is a garage around the corner.\\
A: [A goes to the garage and then meets B again] \\
(*) A: Ok, I got petrol at the garage. Do you think the garage was open? \\
Adapted from \cite[p.311]{grice75}
\medskip \noindent
\end{it}

After acknowledging a contribution at level 4 (which A's \textit{Ok, I got petrol at the garage} clearly does)
it is really hard to go on and ask a
CR about that contribution (A's \textit{Do you think the garage was open?} is a bizarre follow-up --- it could perhaps be interpreted as sarcastic).

Thus our modified proposal for identifying clarifications is the following: \textit{given an utterance U, a subsequent turn is its clarification grounded in modality \emph{m} if it cannot be preceded by positive evidence of understanding of U in \emph{m}}.%
\footnote{For a detailed graphical specification of our recipe, see  Figure 2 in the supplementary material. Notice that the utterances are stored in a stack in Figure 2 because the clarification does not need to be immediately after its source. While an utterance is at the top of the stack it can be clarified, no matter how many turns in between have happened. That way an utterance can be clarified many times.}
Like the earlier version, this implicitly embodies a annotation procedure. Let's see how it works.

%% file: annotation.tex
\section{Grounded clarification annotation} \label{sec-corpus-4}

In this section we evaluate our recipe and the modality-based classification it gives rise to. We do so by using it to annotate a small dataset, the SCARE corpus~\cite{stoia07}. Before delving into the details of the classification, we describe the pragmatic influences that the dialogue participants are under in this dataset.


The SCARE corpus consists of fifteen English spontaneous dialogues situated in
an \emph{instruction giving task}.%
\footnote{The corpus is  available 
at http://slate.cse.ohio-state.edu/quake-corpora/scare/.} The dialogues vary in
length, with a minimum of 400 turns and a maximum of 1500; hence, the dialogues
are \emph{much} longer than other datasets grounded in vision and action where dialogues typically have less than 10 turns on average~\cite{vries:2017, thomason:2019}.
The dialogues were collected
using the \textsc{Quake} environment, a first-person virtual reality game (so
there is
\emph{immediate world validation}). The task consists of a direction giver (DG)
instructing a direction follower (DF) on how to complete several tasks in a
simulated game world. The corpus contains the collected audio and video, as well
as word-aligned transcriptions.  

The DF had no prior knowledge of the world map or tasks and relied on his
partner, the DG, to guide him on completing the tasks (so the DPs have
\emph{asymmetric knowledge of the task}). The DG had a map of the world and a
list of
tasks to complete. The partners
spoke to each other through headset microphones.
As the participants collaborated on the tasks, the DG had instant feedback about
the DF's location in the simulated world, because the game engine displayed the
DF's first person view of the world on both the DG's and DF's computer monitors
(so the DPs \emph{share a view of the task}). Finally, the DPs were
\emph{punished} (they were told they would receive less money for performing the
experiment) if they pressed the wrong buttons or put things in the wrong
cabinets.  

We present a sample interaction from the SCARE corpus. During this dialogue fragment, the dialogue participants were performing one of the tasks of the SCARE experiment specified: \emph{hide the rebreather in cabinet~9}.

The presentation of this dialogue is divided over the two following subsections; the first gives the warm-up necessary for the second. 
Subsection~\ref{subsec-evidence} illustrates how 
positive evidence of understanding is provided, and no examples of CRs are presented here. Subsection~\ref{subsec-CR4}'s goal, on the 
other hand, is to illustrate CRs in different modalities, so here we focus on negative evidence. 

\subsection{Positive evidence} \label{subsec-evidence}

\begin{figure}
    \centering
    
\begin{tabular}{l}
\begin{minipage}{9cm}
\begin{it} 
 DG\dl: see that button straight ahead of you? \\
 DF\dl: \textbf{mhm}\\ 
 DG\dl: hit that one \\
 DF\dl: \textbf{ok}\\
\end{it}
\end{minipage}\\
\begin{minipage}{7cm}
 \centering
 \includegraphics[width=\linewidth]{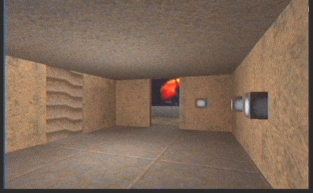} 
\end{minipage} 
\end{tabular}

\caption{Example of the view shared by the dialogue participants and fragment from the SCARE corpus}
\label{positive-evidence}
\end{figure}

At the beginning of this dialogue, the DG is instructing the DF to find
the rebreather. As part of this task, they have to press a button in order to open a door as shown in Figure~\ref{positive-evidence}. The figure shows a dialogue fragment and a screenshot of the shared view when the fragment starts. The turns which provide positive evidence at levels~3 and~4 are shown in boldface. If evidence for proposal is followed by a turn that is not evidence of uptake (of the proposal) then we say that the turn is a CR. 
 
The dialogue fragment reproduced below starts when the DG is trying to get the
DF to press the button that is straight ahead in their
current view; this button opens the cabinet where the rebreather is located. As
part of this project, the DG first makes sure  that the DF identifies this
button using the sub-dialogue constituted by (1) and (2). Once the button is
identified, the short instruction in (3) suffices to convey
the goal of the joint project, namely hitting this button; this is acknowledged at
level~4 in turn~(4) when the DF presses the button.

\subsection{Negative evidence}\label{subsec-CR4}

Now we turn to an extended example, extracted from the SCARE corpus,  of clarification requests at different levels. Between square brackets we indicate forms of non-linguistic communication. The DG utters an instruction in~(1). In turn~(2) the DF makes explicit an obstacle at level~3 that must be solved \emph{before} putting the rebreather in the cabinet, namely  identifying cabinet~9; in doing so he proposes this task. In turn~(3) the DG proposes to identify cabinet~9 by first identifying its location. Turn~(4) is  evidence
of uptake of turn~(3) ---  the DG answers his own question
--- but it is also evidence of the proposal: get back to the starting room.

\begin{it}
\medskip \noindent
DG(1): we have to put it in cabinet nine [pause] \\
DF(2): yeah they're not numbered [laughs] \\
DG(3): [laughs] where is cabinet nine \\
DG(4): it's kinda like back where you started so \\
DF(5): ok so I have to go back through here?\\
DG(6): yeah \\
DF(7): and around the corner?\\
DG(8): right \\
DF(9): and then do I have to go back up the steps?\\
DG(10): yeah \\
DF(11): alright this is where we started \\
DG(12): ok so your left ca-[pause] the left one \\
DF(13): so how do I open it? \\
DF(14): one of the buttons? \\
DG(15): yeah, it's the left one \\ 
DF(16): makes sense \\
DF(17): alright so we put it in cabinet nine \\
\medskip \noindent
\end{it}

Of the 17 turns, 9 were uttered by the DF and 8 by the DG. From the 9 turns by the DF, 5 of them are CRs at level 4 and one at level 3. Turn (2) is a CR of instruction (1).  Turns (5), (7) and (9) are CRs of instruction (4). Utterance (11) shows positive evidence at level 4 of instruction (4) so this instruction cannot be further clarified following the recipe we defined in Section~\ref{CR-id}. Turns (13) and (14) are CRs of utterance (12). The positive evidence at level 4 of instruction (12) is completed by a physical action of the DF in the game world: opening the cabinet by pressing the left button while uttering (16). Finally, turn (17) together with the corresponding physical action are positive evidence at level 4 of instruction (1).   

%% file: analysis.tex
\section{Comparative analysis of clarifications}
\label{sec-to-CR-or-not}

In this section, we identify and discuss a number of
pressures that interact in order to determine the number and type of CRs that occur in dialogue; we also explain why
it makes sense (although it may seem  counter-intuitive at first sight)  that too much uncertainty will tend to \textit{lower} the number of CRs.

The distribution and types of CRs found in a corpus depend on the characteristics of the task that the dialogues in the corpus are addressing. Previous clarification corpus studies~\cite{purver04,rieser05,rodriguez04} have required expensive and detailed annotations by linguists who also evaluated  the quality of the datasets. Purver~\shortcite{purver04} annotates more than 10K turns of the BNC corpus, which contains English dialogue transcriptions of
topics of \emph{general interest} in multiparty dialogue such as meetings. These annotations were used to build a dialogue system that could make and understand relevant clarifications related to different modalities~\cite{purver06}. 
\citep{rieser05} and \cite{rodriguez04} did similar annotations on task-oriented dialogue corpora. 
\cite{rieser05} looked for CRs in a corpus of English
task-oriented human-human dialogue called Communicator. The corpus consists of travel reservation
dialogues between a client a travel agent. The interactions occur by phone;
the
participants do not have a shared view of the task. The corpus comprises 31
dialogues of 67 turns each (on average), from which 4.6\% of the turns are CRs.
12\% of CRs found were classified as level 4 CRs; such as the following: 

\begin{dialogue-lu}
Client: You know what the conference might be downtown Seattle
so I
may have to call you back on that.\\ 
Agent: Okay. Did you want me to wait for the hotel then? 
\end{dialogue-lu}

In this corpus the world validation is informational not physical as in the Bielefeld data that we turn to now.

\cite{rodriguez04} looked for CRs in a corpus of
German task-oriented human-human dialogue called Bielefeld. The dialogues occur in a instruction
giving task for building a model plane. The interactions occur face to face; the
participants have a shared-view of the task. The corpus consists of 22
dialogues, with 180 turns each (on average), from which 5.8\% of the turns are
CRs. 22\% of CRs found were classified as level 4 CRs, such as the following: 
 
\begin{dialogue-lu}
DG: Turn it on. \\
DF: By pushing the red button?
\end{dialogue-lu}

We analyzed the SCARE corpus while watching
the associated videos and we classified the clarification requests according to the levels of
communication using the decision procedure explained in Section~\ref{CR-id}.\footnote{We will release our annotations to the research community upon request.} 
We found that
6.5\% of the turns are CRs. Of these, 65\% belong to  level 4 of
Table~\ref{levels}, and 31\% belong to level 3 (most of them related to
reference resolution). Only 2\% of the CRs were acoustic (level 2) since the
channel used was very reliable, and another 2\% had to do with establishing
contact (level 1). 

The SCARE corpus presents slightly more CRs (at 6.5\%)
than the corpora analyzed in previous work (which reported that 4\%-6\% of the
dialogue turns were CRs). Furthermore, in contrast to the BNC corpus
study~\citep{purver04}, most CRs in the SCARE corpus occurred at level 4. What task characteristics  might have caused the observed differences? 

\begin{table*}[ht]
\begin{center}
\begin{tabular}{|l|c|c|c|c|}
\hline
\textsc{Characteristics} & \textsc{BNC fragment}   & \textsc{Communicator}       & \textsc{Bielefeld}         & \textsc{Scare}               \\
\hline
Task             & Chit-chat & Travel reservation & Building         & Moving              \\
Shared view      & Yes (meetings) & No (on the phone)  & Yes (face-to-face) & Yes (3D game) \\
Participants & More than two & Two & Two & Two \\
World validation & Common ground & Informational      & Physical           & Simulated           \\
Information Flow & Symmetrical & Symmetrical        & Symmetrical        & Asymmetrical        \\
\hline
Total \# turns &   10466   &        2098            & 3962 & 11350 \\
Avg dialogue length  & 30 & 67                 & 180                & 800                 \\
\% of CRs/turns  & 3.1 & 4.6                & 5.8                & 6.8 \\
\hline
\% CRs level 1   & 10 & 3                  & 0                  & 3                   \\
\% CRs level 2   & 31 & 32                 & 12                 & 9                   \\
\% CRs level 3   & 47 & 40                 & 50                 & 32                  \\
\% CRs level 4   & 2 & 12                 & 22                 & 53                  \\
\% CRs other & 10 & 13                 & 16                 & 4                  \\
\hline
\end{tabular}
\caption{Comparing the number of CRs at each level in four corpus studies \label{corpora}}
\end{center}
\end{table*}

We hypothesize that the following six characteristics account for the larger proportion of CRs at level 4 that we find in the SCARE corpus. 
\emph{Task oriented} dialogues (unlike
general interest dialogues) are constrained by the task,
thus the hearer may have a better hypothesis of what the problem is with the source utterance. He also has a clear motivation for asking for clarifications when
the utterance does not fit his model of the task. 
Dialogues  situated in an instruction giving task show an \emph{asymmetry} between the knowledge that the dialogue participants (DPs) have about the task. The Direction Giver (DG) knows how the task has to be done and the Direction Follower (DF) doesn't. Hence, it is to be expected that the DF will have doubts about the task
which (both DPs know) can only be answered by the DG. In symmetric dialogues,
it might not be clear who has what information and then the DPs might not know
who can answer the CRs.  
\emph{Immediate world validation} seems to play a role as well. Dialogues that interleave linguistic actions and informational or physical actions exhibit immediate world validation of the interpretations. If an instruction fails in the world, the DF will ask for clarification. 
When the DPs have a
\emph{shared view} of the task, the DP that is acting on
the world knows
that the other participant is observing him and verifying his actions and then
will try to be sure of what he has to do before doing it. If he is not sure he
will ask. 
\emph{Long dialogues} tend to increase the percentage of clarifications (more than 100 turns) because
DPs prefer to ask questions when they have a good hypothesis to offer. The
longer the interaction, the more background is shared by the
DPs and the easier it will be to come up with a good hypothesis.
Finally, if there are actions in some modality that are \emph{irreversible}, then they
will clarify more until they are sure  of what they have to do.




%% file: discussion.tex
\section{Discussion and objections}
\label{discussion}

Humans switch between clarifications grounded in different modalities seamlessly and we have argued they do so systematically; in effect they do so by following a recipe
for grounding classifications. We obtained this recipe by granting a role to both perceptual and collaborative grounding in
clarification requests. This we did by examining Clark~\shortcite{clark96}'s action ladder of communication and  Ginzburg, Purver and colleagues~\shortcite{ginzburg:2012}'s classification of clarification phenomena, and combining the concept of level
taken from the ladder of communication with 
Gabsdil~\shortcite{gabsdil03}'s  
test for clarification requests. We reframed Clark's downward evidence and upwards completion properties for multimodal interactions. 

This gave us the following:
\textit{given an utterance, a subsequent turn is its clarification grounded in modality \emph{m} if it cannot be preceded by positive evidence of understanding in \emph{m}}. This
provides a unified way to frame clarification mechanisms and their interactions across modalities --- something we view as useful in its own right given the scattered literature
on clarification mechanisms.
However we also suggested that this 
recipe was suitable for learning from data  collected by crowdsourcing. We supported this by examining the
claim that clarifications are rare in dialogue datasets~\cite{ginzburg:2012}, and that current data-hungry algorithms cannot learn them. 
We argued that whether they are rare or not depends on pragmatic factors of the conversation and the modality of the grounded clarification. 
Moreover, along the way we noted a number of practical issues ---
work with \emph{large} dialogues, don't just provide annotators with dialogue fragments, take future conversational history into account when annotating ---
that we think could have an important impact on learnability.

Below we list some possible objections to our proposal. We also include our responses in the hope that this will motivate further debate on these issues in the community. 

\emph{Objection: I still don't have a feel for how much we will gain from this when it comes to a practical, realistic use case; in particular, for an end-to-end system rather than an NLP pipeline.} 

Response: Being able to identify and annotate a turn as a clarification request can help an end-to-end system learn to apply the mechanisms of collaborative grounding to subdialogs, which have rules that differ from modality to modality. 

\emph{Objection: The biggest problem I see is that the distinction of the different levels (which the correct annotation relies on) might not be clear-cut (in particular when considering that crowdsourced annotations usually come from non-experts). I have no idea what quality we get, nor what inter-annotator agreement figures we can expect.}

Response: Our methodology unifies and refines previous methodologies for which inter-annotator agreement has been reported in certain corpora: E.g., .70 for the Bielefeld corpus (Rodríguez and Schlangen, 2004), and .75 for the BNC corpus (Purver, 2004). Our methodology refines Clark (1996) 4-level classification by grounding each level (previously only described by means of examples) to 4 different modalities relevant for situated dialog: socioperception, hearing, vision and movement. This new grounded characterization should improve previous inter-annotator agreement. Using our extended methodology we report a .84 kappa for the SCARE.

\emph{Objection: The corpora that are being investigated are all very domain-specific and relatively small in terms of numbers of dialogues (but with a large average number of turns). This means that even if we were to obtain annotation quality figures, it would still raise the question of what general conclusions we can draw from this.} 

Response: We share this concern; our goal with this paper is to motivate more work in this area.  We believe that this objection actually lends support to our insistence on the importance of a more fine-grained analysis of grounding mechanisms. Our methodology generalizes to domains that ground the communicative intent in the modalities of socioperception, hearing, vision and movement. Examples are robots and virtual assistants, where the dialog partners share a sensible environment. Our argument is that better conceptualizations of clarification subdialogs are needed so that models are able to identify them, distinguish the different types ruled by the different modalities, and learn the structures that govern them.

%% file: conclusions.tex
\section{Conclusions}
\label{sec-conclusions-cap2}


This paper urges the community to address a research gap: how clarification mechanisms can be learned from data. We believe that novel research methodologies which highlight the importance of the role of clarification mechanisms in communicative intent are needed for this. 
So we  presented an annotation methodology, based on a theoretical analysis of clarification requests, which unified a number of previous accounts. 

But to conclude, a different note.
As dialogue systems get better at negotiating meaning with clarifications, future work will need to seriously consider how people relate to conversationally-gifted artificial agents. 
Studies of how users feel when interacting
with dialogue systems~\cite{brave:2005,portela:2017} found that systems can have a psychological impact on users; thus
it will become increasingly important to consider the risks of 
users developing social or emotional bonds with more sophisticated system (thereby affecting their well-being in unforseen ways) and
of users being emotionally manipulated by them. Socioperceptive
dialogue systems  could turn out to have very  sharp teeth indeed. 

%% file: ethical_concers.tex
\section*{Ethical considerations} 
\label{ethics}

In this paper we have not trained machine learning models so we have used negligible computing power. We have not collected a new dataset so we have not used crowdsourcing. The annotation of the SCARE corpus was done by one of the authors and a friend who likes the work and was not economically rewarded. As we noted in the papers conclusion, there are important ethical issues
that future work on this area will need to consider. 
But there are also more immediate discuss ethical considerations and we turn to these now. 

First, the datasets that we use in this paper are described in~\cite{purver04,rodriguez04,rieser05,stoia07}. The dataset in \cite{purver04} contains spoken British English dialogues collected during meetings. The dataset used in~\cite{rieser05} is a fragment of the Carnegie Mellon Communicator Corpus~\cite{bennett:2002}, and is in
American English.  In these dialogues, an experienced travel agent is making reservations for trips that people in the Carnegie Mellon Speech Group were taking in the upcoming months. There is no information to whether the dialogue participants were rewarded or notified about the dataset collection. The dataset in~\cite{rodriguez04} includes dialogues in which one participant gives instructions in German to the other to build a model plane. Finally, the Scare corpus~\cite{stoia07} is an
American English corpus
collected using students at Ohio State University; they were payed to participate in the experiment.  

Future work in this area will need to collect new datasets that reflect the interactions between different types of clarifications in different modalities. Usually such collections are crowdsourced, which raises ethical concerning fair wages and number of hits per day. We would like to encourage the community to value datasets in languages other than English in order to model different strategies for indicating the source of the clarification (prosody, syntactic construction, etc). Last but not least, computing power and carbon footprint should be considered. Machine learning models trained on long multimodal dialogue histories may get very big very fast. We need models that learn to summarize dialogue histories for the sake of the environment and the budget of low-income researchers.

%% file: supplementary.tex
\section{Annotation decision procedure} \label{sec-corpus}

In this section we formalize the classification methodology for clarifications. Our examples here are from the SCARE corpus~\cite{stoia07}. The SCARE corpus consists of fifteen English spontaneous dialogues situated in
an \emph{instruction giving task}.\footnote{The corpus is  available 
in http://slate.cse.ohio-state.edu/quake-corpora/scare/.} The dialogues vary in
length, with a minimum of 400 turns and a maximum of 1500.  

The annotation was performed by two independent annotators with an initial interannotator agreement of .84 kappa. Disagreements were discussed until agreement, a single annotation was obtained. 
We excluded from the annotation  dialogue~1 which has almost no
feedback from the DF because he thought that he was not supposed to speak. 

\begin{figure*}[h!]
 \begin{center}
 \includegraphics[width=1.02\linewidth]{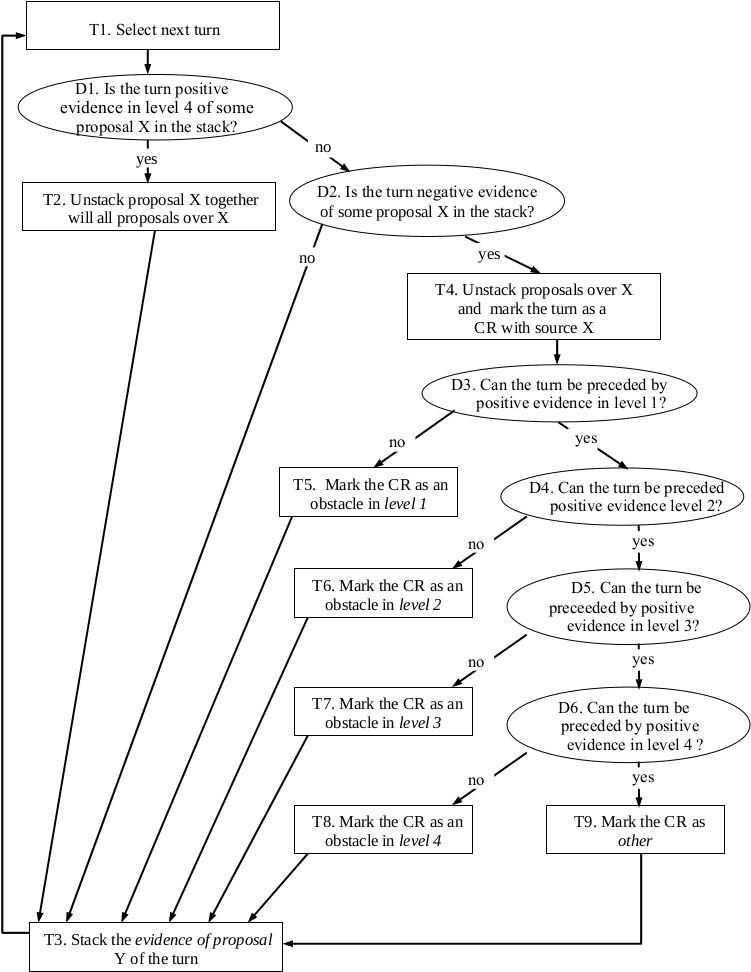} 
 \end{center}
\caption{\label{decision-graph} Decision graph for our recipe. Decision points D1 and D2 decide whether the turn is a CR or
not. Decision points D3 to D6 decide whether the CRs
belong to~\cite{clark96}'s levels 1 to 4 respectively. T9 indicates that the CR is grounded in a modality not represented by levels 1 to 4. }
\end{figure*}

The decision graph used in our ``quasi-systematic'' annotation is
depicted in Figure~\ref{decision-graph}. We call our procedure
``quasi-systematic'' because, while its tasks (depicted in rectangles) are
readily automated, its decision points are not as they require subjective
human judgments. Decision points D1 and D2 decide whether the turn is a CR or
not; new tasks and digressions from the current task answer ``no'' to both
decision points and just stack their evidence of proposal in T3. If the turn is
a CR of a proposal X, T4 unstacks all proposals over X as a result of applying
the downward evidence property of conversations (discussed in
the paper). Intuitively, the turn is taken as an implicit uptake
in level 4 of all the proposals over proposal X (which must be completed
before X can be completed).\footnote{This intuition is in line with
Geurts's preliminary analysis of non-declaratives: if the speaker
did not negotiate the proposals over X, then we can assume that he did not have
problems up-taking them~\citep{geurts10}.}
 Decision points D3 to D6 decide whether the CRs
belong to~\cite{clark96}'s levels 1 to 4 respectively,
with the help of~\cite{rodriguez04}. If a turn can be preceded by positive evidence in level 4 but it still is negative evidence of some proposal made earlier in the dialogue we annotate the CR as \emph{other}. An example of this in dialogue 2 in the SCARE dataset \emph{THERE'S ISN'T REALLY SHORT FOR THERE ARE, IS IT? BUT PEOPLE DO IT ANYWAY} where the DF follower is correcting the DG who said \emph{THERE'S THREE DOORS} earlier. The negative evidence is not related to the modalities relevant for the task at hand but to the language itself. 

